\documentclass{article}
\usepackage[utf8]{inputenc}
\usepackage{authblk}
\usepackage{setspace}
\usepackage[margin=1.25in]{geometry}
\usepackage{graphicx}
\graphicspath{ {./figures/} }
\usepackage{subcaption}
\usepackage{amsmath}
\usepackage[style=nejm, 
citestyle=numeric-comp,
sorting=none]{biblatex}
\title{A soft and lightweight fabric-based pneumatic interface for multimodal fingertip tactile feedback}

\author[1*]{Rui~Chen}
\author[1]{Daniele~Leonardis}
\author[1]{Antonio~Frisoli}

\affil[1]{Institute of Mechanical Intelligence, School of Advanced Studies Sant'Anna (SSSA), 56127 Pisa, Italy.}
\affil[*]{Corresponding author. Email: rui.chen@santannapisa.it}

\date{}

\begin{document}

\maketitle

\begin{abstract}
Wearable fingertip haptic devices are critical for realistic interaction in virtual reality, augmented reality, and teleoperation, yet existing approaches struggle to simultaneously achieve adequate tactile output, low mass, simple fabrication, and untethered portability. Here we show that fabric-based pneumatic actuation can address this gap. Our device comprises four pneumatic chambers fabricated from thermoplastic polyurethane-coated fabric via computer numerical control heat-sealing, yielding a soft, conformable interface weighing 2.1\,g that operates untethered with a wrist-mounted control unit. Mechanical and dynamic characterization confirms that the fabric actuators produce sufficient force, displacement, and bandwidth for fingertip tactile rendering. A psychophysical study with 15 participants demonstrates classification accuracy exceeding 90\% across three distinct tactile modes---contact configuration, directional sliding, and vibrotactile frequency. These findings establish fabric-based pneumatic actuation as a viable technology route for lightweight, low-cost, and multimodal fingertip haptic interfaces.
\end{abstract}

% ============================================================
\section{Introduction}
% ============================================================

% --- Para 1: Broad significance ---
Touch is fundamental to how humans interact with the physical world. During object manipulation, the fingertips, among the most densely innervated regions of the body, supply a continuous stream of tactile information, including contact force, surface geometry, compliance, texture, and slip, that vision and audition alone cannot provide~\cite{Pacchierotti2017_review_finger_hand,fleck2025multi_sensory_Review}. Reproducing this rich sensory experience at the fingertip is therefore a central challenge for immersive technologies. In virtual and augmented reality (VR/AR) and teleoperation, the absence of reliable fingertip haptic feedback remains a significant barrier to realistic interaction, limiting both task performance and the user's sense of physical presence~\cite{frisoli2024VR_review,shi2024sensing_feedback_VR_Review,pacchierotti2023cutaneous_teleoperation}.

% --- Para 2: Landscape of existing approaches ---
A broad range of actuation technologies has been explored for wearable fingertip haptic devices, including electromagnetic motors~\cite{kim2025motor_cable_3DOf_touch,Xu2025Motor_3DOF}, piezoelectric actuators~\cite{carluccio2023piezoelectric,jiang2024multimodal_piezoelectric}, shape memory alloys~\cite{chen2026SMA,baba2024pin_array_SMA}, dielectric elastomer actuators~\cite{guo2024_DEA_Skin,youn2025skin_DEA,bai2021elastomeric_VR_review}, pneumatic actuators~\cite{du2024haptiknit,mazzotta2025_thermo_pneumatic,Rui2026_temp_pouch,sonar2021soft_touch_pneumatic,kang2025Pneumatic_ring,stanley2015Jaming}, electrohydraulic actuators~\cite{hartcher2023electrohydraulic_16,shao2025HAESL_single_Electrohydraulic,sanchez2024_HAESL_single_Wrist_electrohydraulic}, and electrotactile stimulation~\cite{yao2024_electrotactile,lin2022super_resolution_electrotactile,huang2023skin_multimodal_electrotactile}, targeting stimuli that span force, shape, texture, and temperature~\cite{chen2026review_materails_structure_haptic}. A common strategy for improving spatial resolution has been to distribute dense actuator arrays across the fingertip contact area, using configurations of dielectric elastomer actuators~\cite{guo2024_DEA_Skin,youn2025skin_DEA,youn2021wearable_DEA}, electroosmotic pumps~\cite{shen2023_electroosmotic_pump,shultz2023_electroosmotic_pumps}, hydraulically amplified self-healing electrostatic actuators~\cite{shao2025HAESL_single_Electrohydraulic,sanchez2024_HAESL_single_Wrist_electrohydraulic,hartcher2023electrohydraulic_16,leroy2020_Multimode_HESAL,johnson2023actuation_sensing_Hasel}, voice coils~\cite{tursynbek2025_PinArray_Voice_Coil,li2022touchIOT_coil,grasso2023HAXEL_arrays}, electrotactile electrodes~\cite{lin2022super_resolution_electrotactile,yao2024_electrotactile,huang2023skin_multimodal_electrotactile}, or pneumatically driven pins~\cite{ujitoko2020_pin_pneumatic,shan2025transparent_pneumatic_high_resolution_49,abad2024_Pneumatic_14_Vibration,morita2023Pneumatic_vaccum_16,wu2024_pin_motor,Sarakoglou2012_Cable_Pin}. 
In the above approaches, device design requires placement of an actuating mechanism close to the fingerpad area, hence scaling mass, complexity and encumbrance of the device with the number of actuated area or degrees of freedom. 
As more actuators are packed into the limited fingertip area, each must shrink, reducing force output and displacement, while the associated electronics, pneumatic supply lines, or high-voltage circuitry add substantial mass and often necessitate tethered operation~\cite{son2026soft_interface_review_position}. As a result, simultaneously achieving adequate contact force, meaningful surface deformation, practical wearability, simple fabrication, and untethered portability within a single device remains an open challenge.

% --- Para 3: Why fabric-based pneumatic ---
Fabric-based pneumatic actuators---constructed from thermoplastic polyurethane (TPU)-coated textiles and formed by heat-sealing---offer a qualitatively different set of design characteristics~\cite{nguyen2020design_modeling}. The actuator material is intrinsically lightweight, soft, and conformable; fabrication via CNC heat-sealing requires no cleanroom facilities, dedicated moulds, or high-voltage equipment, and produces arbitrary planar chamber geometries with high reproducibility at low cost~\cite{chen2026_GPAs,Goshtasbi_2025_WIld_CNC,gohlke2023wireshape_CNC,feng2023_low_cost}. These attributes have enabled successful deployment in other domains of soft robotics, particularly soft wearable robotics~\cite{Rui_LPPAMs,Rui2026_temp_pouch}. However, whether fabric-based pneumatic actuation can be designed to deliver perceptually meaningful tactile feedback at the fingertip has not been systematically investigated. Here actuator dimensions are constrained to the sub-centimetre scale, force output must be sufficient to elicit reliable tactile sensation, and dynamic bandwidth must support multiple stimulation modalities. This gap motivates the present work.

% --- Para 4: What we do ---
Here we present a fingertip haptic interface comprising four independently addressable fabric-based pneumatic chambers in a 2$\times$2 arrangement, fabricated from TPU-coated fabric via CNC heat-sealing (Fig.~\ref{Fig-M-Conceptual}). A fully portable pneumatic control box worn on the wrist enables untethered operation. By selectively activating chambers in distinct spatial and temporal patterns, the device supports three perceptually different stimulation modes (contact configuration, directional sliding, and vibrotactile frequency) from a single lightweight platform.

% --- Para 5: Preview of results ---
We report systematic characterization of the fabric actuator's mechanical and dynamic properties, quantification of the spatial pressure distribution at the finger--actuator interface, results of a psychophysical study with 15 participants evaluating perceptual discriminability across all three stimulation modes, and a representative VR demonstration of real-time haptic rendering. Experimental results establish that fabric-based pneumatic actuation can provide perceptually reliable multimodal tactile feedback at the fingertip within an ultralight and portable form factor, supporting this technology as a viable and practical approach for wearable haptic interfaces in VR/AR and teleoperation applications.

% ============================================================
\section*{Results}
% ============================================================

% --------------------------------------------------------
\subsection*{Device Design and Fabrication}
% --------------------------------------------------------

\begin{figure}[htbp]
    \centering
    \includegraphics[width=1\textwidth]{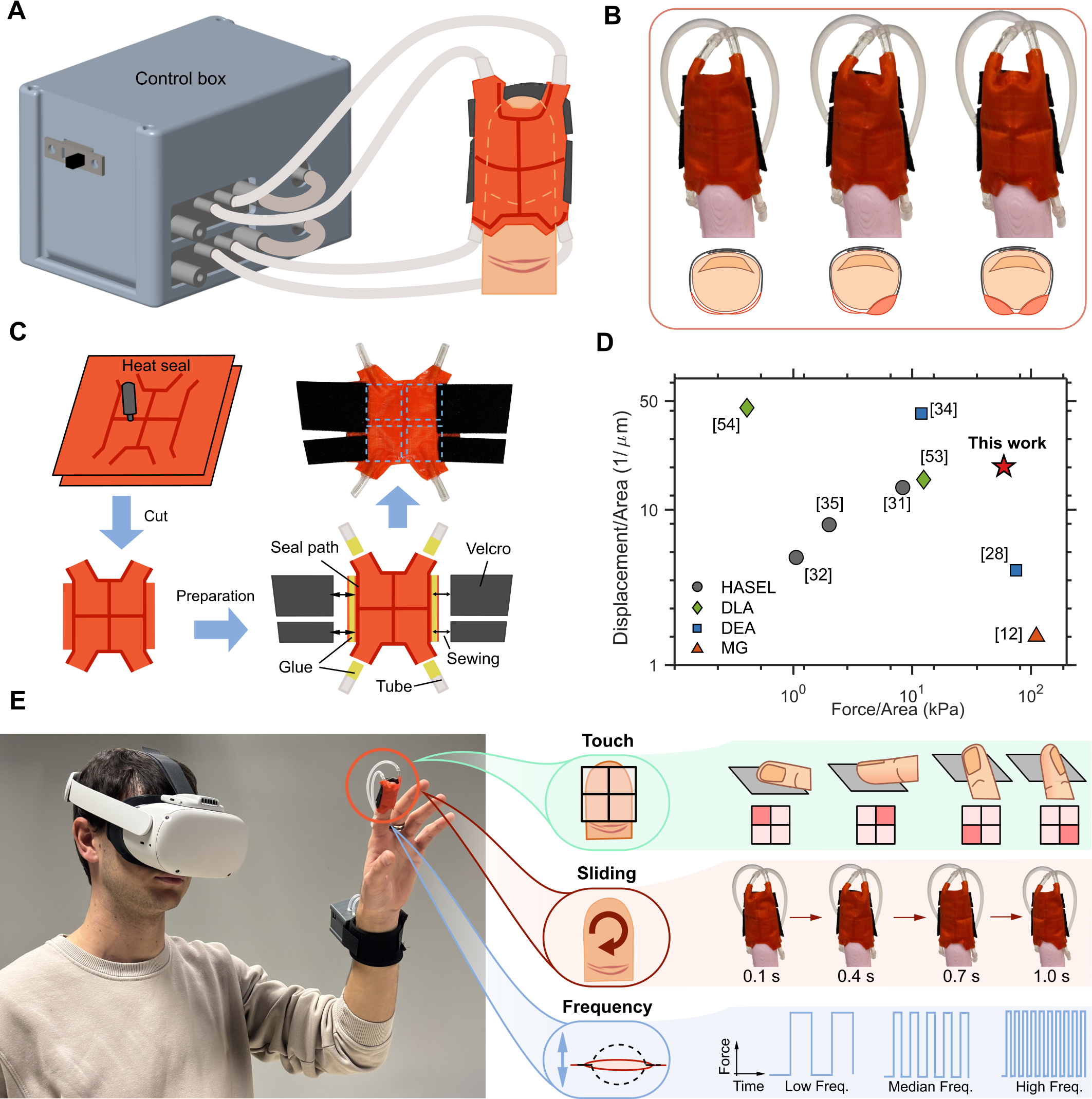}
    \caption{\textbf{Overview of the fabric-based pneumatic fingertip haptic interface.}
    \textbf{(A)}~System architecture showing the wrist-mounted control box, pneumatic tubing, and fingertip haptic interface.
    \textbf{(B)}~The 2$\times$2 chamber configuration enabling independently addressable spatiotemporal actuation patterns.
    \textbf{(C)}~Fabrication process via CNC heat-sealing and assembly of the soft wearable interface.
    \textbf{(D)}~Performance comparison against prior fingertip haptic devices in normalised force output and displacement.
    \textbf{(E)}~Three stimulation modes supported by the device: contact configuration, directional sliding, and vibrotactile feedback.
    }
    \label{Fig-M-Conceptual}
\end{figure}

The haptic device comprises three principal components: a portable control box mounted on the wrist, a wearable fingertip haptic interface, and four pneumatic tubes interconnecting them (Fig.~\ref{Fig-M-Conceptual}A). The haptic interface incorporates four independently addressable pneumatic chambers (pouch motors~\cite{niiyama2015pouch_motor}) arranged in a 2$\times$2 configuration. The choice of four chambers reflects the practical constraints of the current system: at the fingertip scale, routing individual air supply tubes to each chamber without mutual interference becomes increasingly difficult as channel count rises, and the miniaturised portable control box accommodates four independent pneumatic channels. This configuration is not asserted as optimal; rather, it represents the channel budget achievable within the present system integration constraints (Fig.~\ref{Fig-M-Conceptual}B). Through selective and sequential chamber activation, this arrangement enables programmable spatiotemporal actuation patterns that deliver spatially distributed tactile stimulation to the fingertip.

The pneumatic chambers are constructed from TPU-coated ripstop nylon fabric via CNC heat-sealing, a process in which a thermal sealing head mounted on a repurposed desktop 3D printer platform traces programmable paths to bond two fabric layers, forming airtight chambers of arbitrary planar geometry with high repeatability and minimal manual intervention~\cite{Rui_2025_customized_glove,nguyen2020design_CNC,chen2026_GPAs,Goshtasbi_2025_WIld_CNC,gohlke2023wireshape_CNC}. Compared with rigid, elastomeric, or piezoelectric alternatives, TPU-coated fabric is lightweight, intrinsically soft, and suitable for low-cost processing without specialized equipment (Fig.~\ref{Fig-M-Conceptual}C)~\cite{feng2023_low_cost,Rui_LPPAMs}. A notable characteristic of this fabrication approach is that geometric complexity adds negligible manufacturing burden: changing the number, shape, or arrangement of chambers requires only a modification to the CNC sealing path, with no additional tooling, moulds, or assembly steps. By bonding the four-chamber actuator assembly to air supply tubes and a Velcro retention strap, a conformable wearable interface weighing only 2.1\,g is realised. Figure~\ref{Fig-M-Conceptual}D positions the present device relative to prior portable fingertip haptic interfaces in the normalised force--displacement design space, where force and displacement are divided by actuator area to facilitate comparison across devices of differing footprint (Supplementary Table 2). Through selective and sequential chamber activation, the device supports three distinct stimulation modes---contact configuration, directional sliding, and vibrotactile stimulation---enabling versatile fingertip feedback from a single platform (Fig.~\ref{Fig-M-Conceptual}E).

% --------------------------------------------------------
\subsection*{Actuator Characterisation}
% --------------------------------------------------------

\begin{figure}[htbp]
    \centering
    \includegraphics[width=1\textwidth]{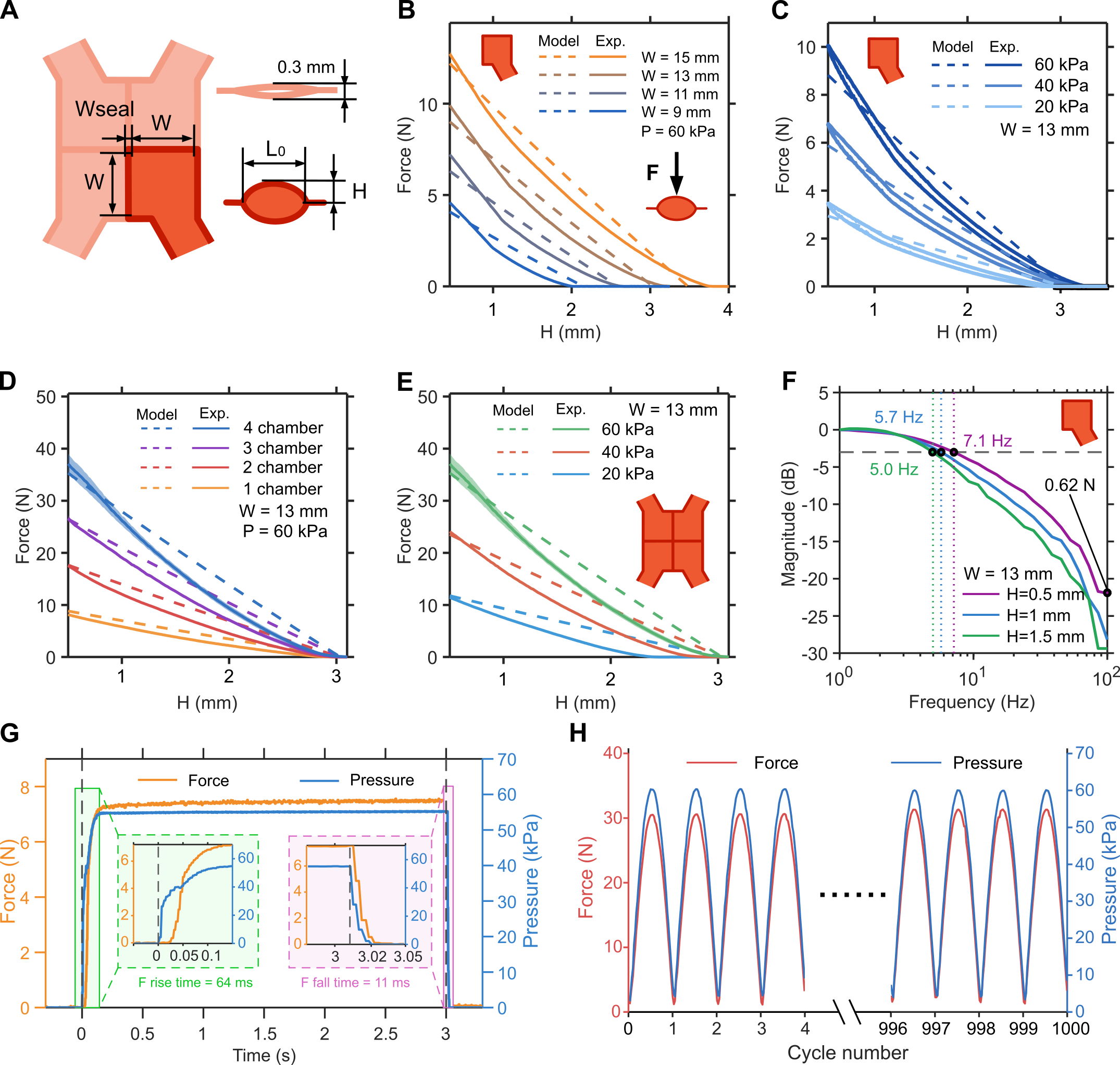}
    \caption{\textbf{Mechanical and dynamic characterisation of the fabric actuator.}
    \textbf{(A)}~Geometric parameters of the haptic actuator.
    \textbf{(B)}~Modelled and experimental force--displacement results across actuator widths $W$.
    \textbf{(C)}~Pressure--hysteresis curves of the single-chamber actuator.
    \textbf{(D)}~Force output as a function of chamber number.
    \textbf{(E)}~Force output of the four-chamber actuator across three pressure levels.
    \textbf{(F)}~Bode plot demonstrating a $-$3\,dB bandwidth of 7.1\,Hz.
    \textbf{(G)}~Step response showing 64\,ms rise time and 11\,ms fall time.
    \textbf{(H)}~Force and pressure output stability over 1{,}000 actuation cycles.}
    \label{Fig-M-Actuator}
\end{figure}

Having established the device design, we next characterised the mechanical and dynamic performance of the fabric actuator to quantify its rendering capability and delineate the operating envelope of this actuator technology at the fingertip scale. Key geometric parameters are defined in Fig.~\ref{Fig-M-Actuator}A, and a simplified analytical model was derived to predict actuator behaviour across design variants (see Supplementary Notes and Supplementary Figure 1A).

\textbf{Force--displacement performance.}
Model predictions closely matched experimental measurements of the force--height relationship across actuator widths $W$ (Fig.~\ref{Fig-M-Actuator}B). Force decreases nonlinearly with increasing actuation height $H$, while both peak force and maximum displacement increase monotonically with $W$, reaching approximately 12\,N and 3.7\,mm at the largest tested width. For a fixed width of $W = 13$\,mm, the actuator exhibited minimal hysteresis across the tested pressure range, sustaining up to 10\,N force and 3.2\,mm displacement (Fig.~\ref{Fig-M-Actuator}C). These results confirm that each chamber produces well-defined and repeatable force and displacement outputs across input pressures, providing a physical basis for spatially selective tactile stimulation through chamber selection. Activating multiple chambers in parallel additively increases output force: a four-chamber configuration achieves up to 36\,N, with force scaling proportionally with input pressure (Fig.~\ref{Fig-M-Actuator}D--E). It should be noted that these measurements were obtained against a rigid constraint; in practice, the compliance of the fabric actuator membrane and human fingertip tissue will redistribute contact forces, and the effective force delivered to the skin is expected to be lower than the bench-measured values.

\textbf{Dynamic response.}
Frequency response characterisation revealed a $-$3\,dB bandwidth of 7.1\,Hz at a constrained displacement of 0.5\,mm (Fig.~\ref{Fig-M-Actuator}F), with bandwidth increasing as displacement amplitude decreases. Even at 100\,Hz, a residual force amplitude of 0.62\,N was measured, which exceeds the perceptual threshold for vibrotactile stimulation at the fingertip~\cite{patel2026Touch_review}. Step response testing demonstrated a rise time (10\%--90\%) of 64\,ms and a fall time (90\%--10\%) of 11\,ms (Fig.~\ref{Fig-M-Actuator}G), reflecting the low impedance of the miniaturised valve--pump assembly. These dynamic characteristics are relevant to all three stimulation modes: the 7.1\,Hz bandwidth supports the temporal patterns used for contact configuration and sliding rendering, while the rapid on/off switching enables vibrotactile stimulation at frequencies where force amplitude is attenuated but remains perceptible.

\textbf{Durability.}
Long-term reliability was assessed via cyclic testing under sinusoidal pressure input (0--60\,kPa). Force output remained stable throughout 1{,}000 cycles (Fig.~\ref{Fig-M-Actuator}H), confirming mechanical durability suitable for extended wearable use.

% --------------------------------------------------------
\subsection*{Spatial Pressure Distribution}
% --------------------------------------------------------

The actuator characterisation established the force and dynamic capabilities of individual chambers. To verify that these capabilities translate into spatially distinguishable tactile stimulation at the fingertip, we next quantified the spatial pressure distribution at the finger--actuator interface.

\begin{figure}[htbp]
    \centering
    \includegraphics[width=0.9\textwidth]{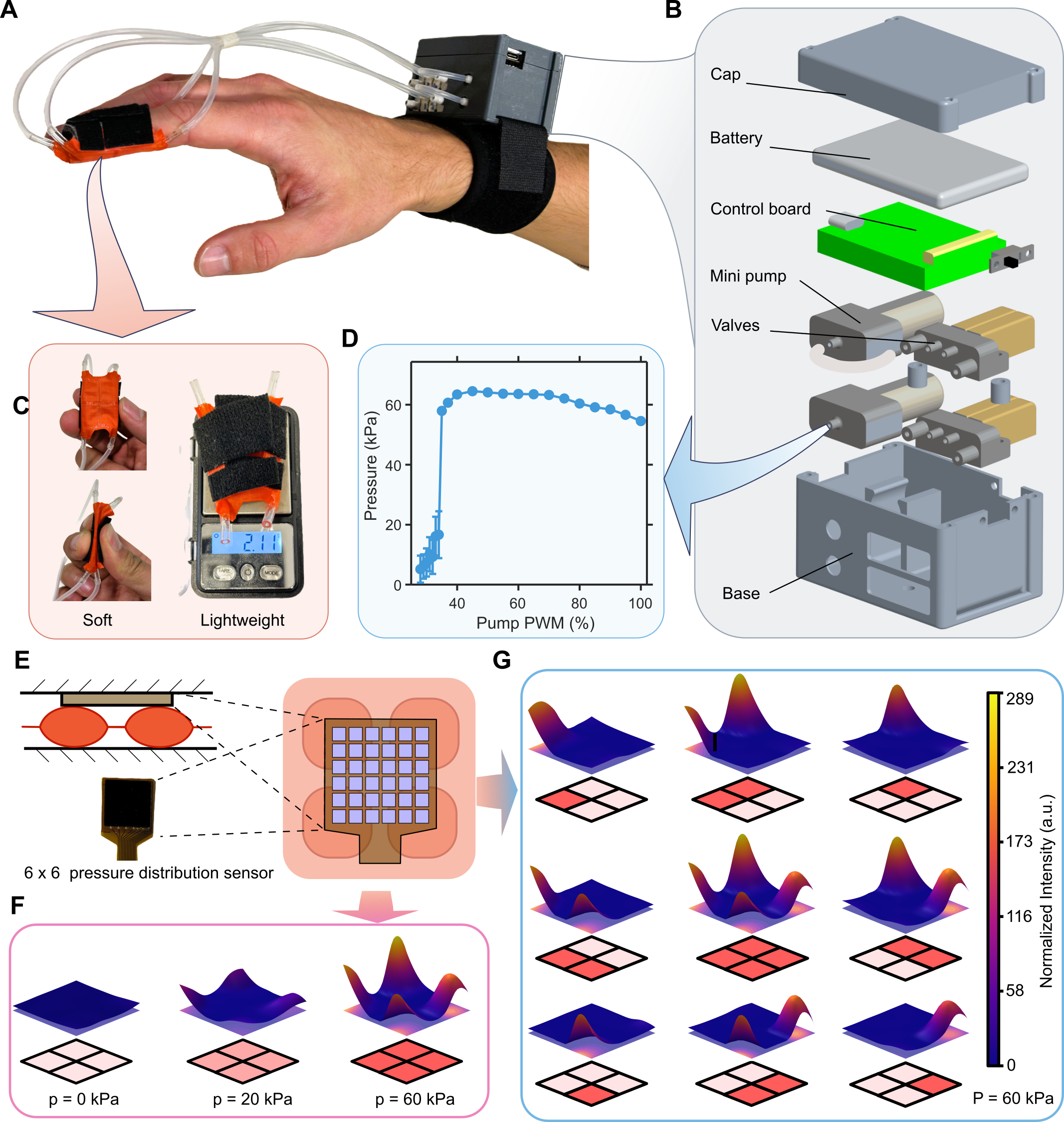}
    \caption{\textbf{Portable haptic device and spatial pressure distribution.}
    \textbf{(A)}~Haptic device worn on the index finger.
    \textbf{(B)}~Exploded view of the control box showing internal components.
    \textbf{(C)}~The soft haptic interface weighs only 2.1\,g.
    \textbf{(D)}~Pump pressure output as a function of PWM duty cycle.
    \textbf{(E)}~Experimental setup for pressure distribution measurement.
    \textbf{(F)}~Pressure distribution under varying actuation pressure levels.
    \textbf{(G)}~Pressure distribution across different chamber actuation patterns.}
    \label{Fig-M-Haptic-Device}
\end{figure}

Integration of the haptic interface with a custom miniaturized pneumatic control box yielded a fully portable, self-contained system (Fig.~\ref{Fig-M-Haptic-Device}A--B, and Supplementary Figure 2A--B). The soft haptic interface weighs 2.1\,g, while the complete control box, including two micro-pumps, four solenoid valves, an ESP32-C3 microcontroller, and a lithium-ion battery, weighs 183.14\,g (Fig.~\ref{Fig-M-Haptic-Device}C). The two onboard micro-pumps deliver up to 64\,kPa, with output partially regulated via pulse-width modulation (PWM), at a combined theoretical flow rate of 1.6\,L/min (Fig.~\ref{Fig-M-Haptic-Device}D).

Spatial pressure distribution was quantified using a 6$\times$6 piezoresistive tactile sensor array (36 sensing elements) placed between the haptic interface and a rigid constraint (Fig.~\ref{Fig-M-Haptic-Device}E). Two key observations emerged. First, increasing actuation pressure produced proportionally greater contact force across the sensor array (Fig.~\ref{Fig-M-Haptic-Device}F). Second, and crucial for the intended application, varying the chamber actuation pattern systematically shifted the spatial distribution of contact pressure across distinct fingertip regions (Fig.~\ref{Fig-M-Haptic-Device}G and Supplementary Figure 1E). These results confirm that selective chamber activation generates spatially configurable pressure profiles at the fingertip, representing the experimental basis for the three stimulation modes evaluated in the subsequent perceptual study.

% --------------------------------------------------------
\subsection*{Human Perception Study}
% --------------------------------------------------------

To evaluate whether the haptic interface delivers perceptually meaningful feedback, we conducted a psychophysical study comprising three forced-choice identification tasks, each corresponding to one stimulation mode. Fifteen healthy adult participants (4 female, 11 male; age $30.4 \pm 3.7$\,years) were enrolled. Auditory isolation was ensured via noise-cancelling headphones delivering white noise. Stimuli were presented in pseudorandom order, and participants reported their judgement verbally or by pointing to a visual reference sheet, with no time constraint imposed (Supplementary Figure 2D). Each task was accompanied by a corresponding VR demonstration scenario to illustrate the interactive context in which the stimulation mode would be deployed.

\begin{figure}[htbp]
    \centering
    \includegraphics[width=1\textwidth]{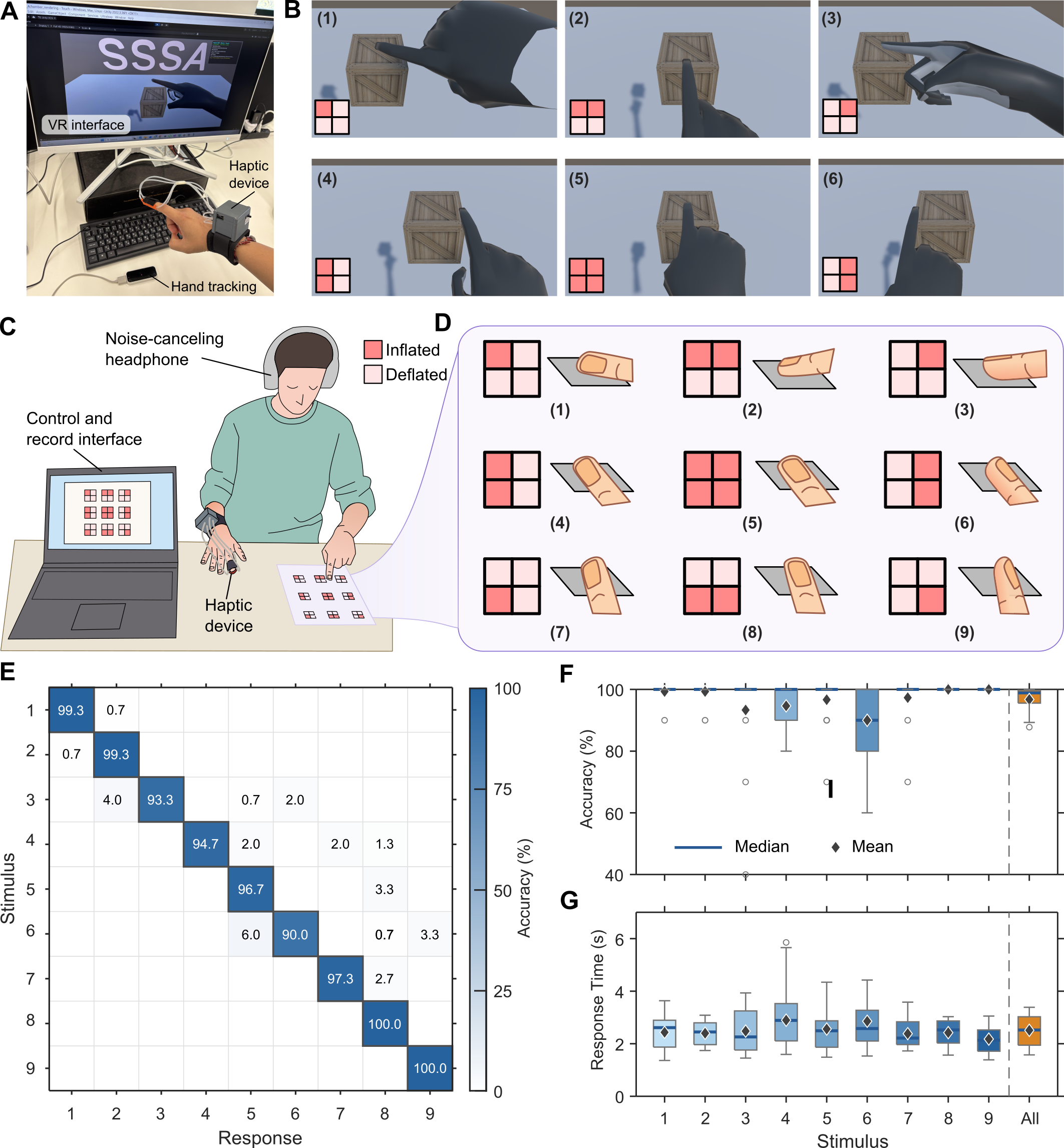}
    \caption{\textbf{Human perception study: contact configuration discrimination task.}
    \textbf{(A)}~VR environment used for touch simulation.
    \textbf{(B)}~Distinct actuation patterns generated under different contact configurations.
    \textbf{(C)}~Experimental setup for the user study.
    \textbf{(D)}~Nine actuation patterns corresponding to different contact configurations.
    \textbf{(E)}~Confusion matrix illustrating an overall classification accuracy exceeding 94\%.
    \textbf{(F, G)}~Accuracy and response time distributions across different stimulus patterns.}
    \label{Fig-M-Subject1}
\end{figure}

\textbf{Contact configuration.}
In the accompanying VR scenario, a virtual cube was placed on a platform, and the user's hand, tracked by Leap Motion, could freely explore the object (Fig.~\ref{Fig-M-Subject1}A and Supplementary Video~1). When the fingertip contacted the cube at different angles or along its edges, only a subset of the four quadrants penetrated the virtual surface; the corresponding chambers were actuated while the remaining chambers stayed idle. This spatially selective feedback conveyed the contact configuration (specifically, which region of the fingerpad was in contact) through differential pressure on the fingerpad (Fig.~\ref{Fig-M-Subject1}B). In the discrimination task, nine distinct actuation patterns encoding different contact configurations were presented (Fig.~\ref{Fig-M-Subject1}C--D). Participants achieved an overall classification accuracy of $96.74 \pm 3.79\%$, with accuracy exceeding 90\% for all nine patterns and reaching 100\% for patterns 8 and 9 (Fig.~\ref{Fig-M-Subject1}E--G). Mean response time was $2.51 \pm 0.59$\,s, measured from stimulus onset to response recording.

\begin{figure}[htbp]
    \centering
    \includegraphics[width=1\textwidth]{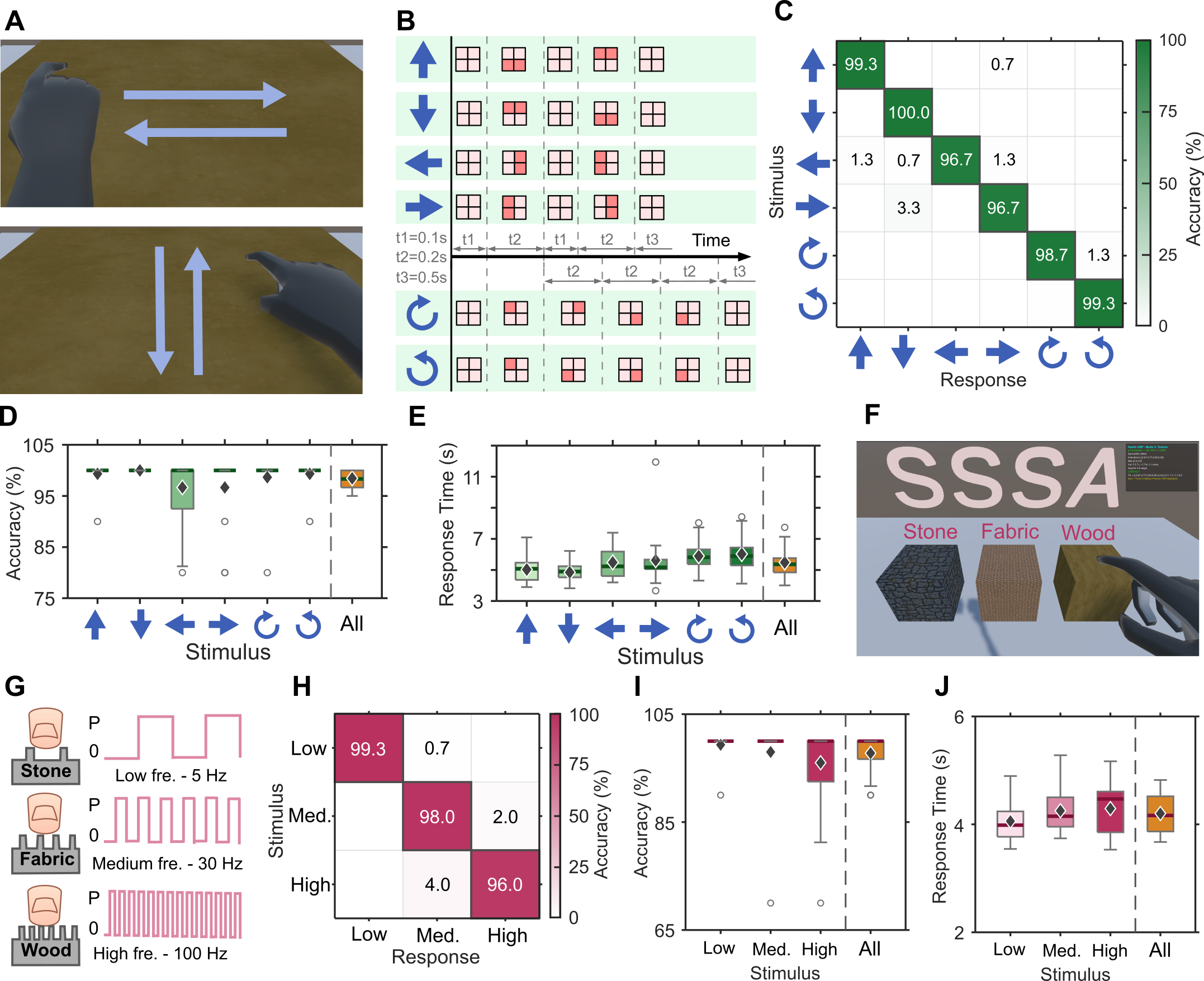}
    \caption{\textbf{Human perception study: sliding and vibrotactile discrimination tasks.}
    \textbf{(A)}~VR environment for the sliding task.
    \textbf{(B)}~Schematic of six directional sliding patterns.
    \textbf{(C)}~Confusion matrix for the sliding task, indicating an overall classification accuracy above 96\%.
    \textbf{(D, E)}~Accuracy and response time distributions across sliding stimuli.
    \textbf{(F)}~VR environment for vibrotactile-based texture representation.
    \textbf{(G)}~Schematic of three vibrotactile frequency conditions.
    \textbf{(H)}~Confusion matrix for the vibrotactile task, indicating an overall classification accuracy above 98\%.
    \textbf{(I, J)}~Accuracy and response time distributions across frequency conditions.}
    \label{Fig-M-Subject2}
\end{figure}

\textbf{Directional sliding.}
In the second VR scenario, a large flat surface was presented in the virtual environment, and the user's fingertip could slide freely across it (Fig.~\ref{Fig-M-Subject2}A and Supplementary Video~2). When the fingertip was in contact and the principal direction of fingertip velocity exceeded a predefined threshold, a sequential chamber actuation pattern was delivered to convey the perceived sliding direction (up, down, left, or right) as well as clockwise or counterclockwise rotation (Fig.~\ref{Fig-M-Subject2}B). Six motion sequences were presented in the discrimination task. Participants achieved an overall accuracy of $98.44 \pm 1.72\%$, with accuracy exceeding 96\% for all six patterns and reaching 100\% for the downward direction (Fig.~\ref{Fig-M-Subject2}C--E). Response times were longer for this mode ($5.48 \pm 0.91$\,s), consistent with the additional temporal integration required to perceive sequential actuation patterns.

\textbf{Vibrotactile frequency.}
In the third VR scenario, three virtual objects with distinct surface materials---rough stone, medium-rough fabric, and smooth wood---were arranged on a platform (Fig.~\ref{Fig-M-Subject2}E and Supplementary Video~3). When the fingertip contacted an object, the chambers in contact were driven at a material-specific vibration frequency: 5\,Hz for stone, 30\,Hz for fabric, and 100\,Hz for wood. Participants achieved an overall accuracy of $97.78 \pm 3.71\%$, with accuracy exceeding 96\% at each frequency (Fig.~\ref{Fig-M-Subject2}F--G). Mean response time was $4.20 \pm 0.39$\,s.

\textbf{Effects of anthropometric factors.}
We additionally examined the effects of finger circumference and biological sex on accuracy and response time across all three tasks. No statistically significant differences were observed (Supplementary Figure~3). However, the limited sample size restricts the statistical power to detect small between-group effects, and these findings should be interpreted with caution.

\textbf{Summary.}
Across all three stimulation modes, the device achieved classification accuracy exceeding 90\%, well above the respective chance levels, with response times within ranges compatible with interactive haptic feedback. These results demonstrate that fabric-based pneumatic actuation at the fingertip scale can support perceptually reliable discrimination of spatial contact configuration, motion direction, and vibrotactile frequency, using the present four-chamber configuration in a wearable interface weighing approximately two grams.

% ============================================================
\section*{Discussion}
% ============================================================

This work investigated whether fabric-based pneumatic actuation can deliver perceptually meaningful tactile feedback when scaled down to the size of a fingertip haptic device. The study was motivated by the potential of fabric-based pneumatic actuators shown in other soft robotics domains. The results indicate that across three stimulation modes (contact configuration, directional sliding, and vibrotactile frequency), the approach obtains classification accuracy exceeded 90\% in all tasks, achieved within an ultralight (2.1\,g) and fully portable device.

% --- Paragraph 2: Significance of perceptual results ---
Several aspects of the perceptual findings merit discussion. The high accuracy in contact configuration discrimination ($96.74 \pm 3.79\%$ across nine patterns) indicates that four independently controlled chambers in a 2$\times$2 arrangement generate sufficient spatial contrast at the fingertip for participants to reliably identify distinct contact configurations, which can in turn be used to provide information of angled surface contact and partial edge contact. Directional sliding achieved the highest accuracy among the three tasks ($98.44 \pm 1.72\%$), suggesting that sequential chamber activation is a particularly effective encoding strategy for conveying motion direction, a cue relevant to slip detection and surface exploration during manipulation. Vibrotactile frequency discrimination across three bands (5, 30, and 100\,Hz) was similarly reliable ($97.78 \pm 3.71\%$); these frequencies correspond broadly to the sensitivity ranges of distinct mechano-receptor populations involved in tactile frequency perception, suggesting compatibility between the device's output characteristics and the human somatosensory system. It is important to note, however, that these results reflect performance on a forced-choice identification task under controlled conditions. Whether the same level of perceptual reliability translates to improved task performance in ecologically valid manipulation or interaction scenarios remains to be established in future work.

% --- Paragraph 3: Comparison with prior work ---
The fabric-based pneumatic approach achieves a combination of features that include ultralight mass, multi-chamber and multi-modal actuation at the fingertip scale, clearly perceivable stimuli, portable operation, that, to our knowledge, have not been jointly demonstrated in a single device. This does not imply superiority over alternative technologies in any single performance dimension; devices based on dense actuator arrays may offer higher spatial resolution~\cite{guo2024_DEA_Skin,tursynbek2025_PinArray_Voice_Coil,ujitoko2020_pin_pneumatic,jang2024Strain_big_dielectric_liquid,han2020haptic_dielectric_fluid_transducers}, while piezoelectric or electromagnetic actuators may provide greater bandwidth~\cite{jiang2024multimodal_piezoelectric,li2022touchIOT_coil,grasso2023HAXEL_arrays}. Rather, the present work demonstrates that fabric-based pneumatic actuation occupies a distinct and previously underexplored region of the design space for fingertip haptics, where wearability, fabrication simplicity, and cost are prioritized alongside adequate perceptual performance.

% --- Paragraph 4: Fabrication characteristics ---
The fabrication approach offers several practical advantages that are worth to highlight. CNC heat-sealing produces arbitrary planar chamber geometries from commodity textile materials, requiring no cleanroom facilities, specialised moulds, lithographic patterning, or high-voltage equipment. Fabrication time for a single device is on the order of minutes, and the process is inherently scalable and customisable: modifying the chamber geometry (number, shape, or arrangement) requires only a change to the CNC toolpath, with no additional tooling or assembly steps. These characteristics lower the barrier to adoption and iteration, and may be particularly relevant for applications requiring user-specific customization.

% --- Paragraph 5: System integration constraints and future directions ---
The present device configuration, featuring four chambers in a fixed 2$\times$2 arrangement, reflects the constraints of the current system integration rather than a claim about the optimal actuator count. At the fingertip scale, routing individual air supply tubes to each chamber without mutual interference becomes progressively more difficult as the number of channels increases, and the miniaturized portable control box currently accommodates four independent pneumatic channels with binary on/off valve switching. Notably, the fabrication process itself imposes essentially no constraint on chamber count or geometry: the CNC heat-sealing toolpath can readily produce higher channel counts. The practical bottleneck for increasing spatial resolution lies in the pneumatic routing at the fingertip scale and in the miniaturization of multi-channel portable control electronics. Future work aimed at overcoming this constraint (i.e. through manifold-integrated tubing, micro electromechanical valves, or multiplexed pneumatic control) could substantially extend the capability of this platform while preserving its fabrication advantages.

% --- Paragraph 6: Additional limitations ---
Several additional limitations have to be acknowledged. First, the portable pneumatic control system currently supports only binary on/off valve switching; incorporating proportional pressure regulation and closed-loop control within each channel would enable independent pressure modulation, expanding rendering capabilities beyond the present spatially selective activation scheme. Second, the temporal parameters used in the sliding and vibrotactile tasks, including inter-chamber intervals and stimulus frequencies, were selected based on pilot experiments; systematic psychophysical characterization of perceptual thresholds across a broader parameter space would further inform optimal stimulus design. Third, the user study enrolled 15 participants, which was sufficient to demonstrate high classification accuracy with adequate statistical power for the primary measures, but limits the ability to detect small between-group effects or to draw strong conclusions about individual difference factors. Larger and more diverse participant samples will strengthen these conclusions. Fourth, the perceptual study evaluated pattern discriminability under controlled conditions; assessment of the device's impact on task performance in realistic VR/AR or teleoperation scenarios remains an important direction for future investigation.

% --- Paragraph 7: Outlook ---
Looking forward, the lightweight, untethered, and low-cost nature of the fabric-based platform positions it as a practical building block for multi-finger or full-hand haptic systems. The device architecture is well suited for extension towards closed-loop haptic rendering: by integrating real-time contact force and geometry estimation from sensor data, the system could faithfully reproduce fingertip contact events during virtual or remote interaction, closing the loop between digital environments and tactile perception. We hope the systematic characterization and perceptual validation reported here, as well the limitations of the device discussed in this section, will prove useful for researchers exploring fabric-based pneumatic actuation as a technology route for wearable haptic interfaces.
%More broadly, we believe that the systematic characterization and perceptual validation reported here will serve as a reference point for researchers exploring fabric-based pneumatic actuation as a technology route for wearable haptic interfaces.

% ============================================================
\section*{Materials and Methods}
% ============================================================

\subsection*{Materials and Fabrication}

TPU-coated ripstop nylon fabric (Adventure Expert) was used throughout. Chamber membranes were fabricated from 40D ripstop nylon TPU fabric. Actuator chambers were formed via CNC heat-sealing using a thermal sealing head mounted on a repurposed desktop 3D printer platform (Anycubic Mega), enabling programmable sealing of arbitrary planar geometries with high repeatability. Air supply tubes were bonded to each chamber inlet port using cyanoacrylate adhesive.

\subsection*{Pneumatic Control}

The portable control box architecture is illustrated in Supplementary Figure~2C. Two miniaturised diaphragm pumps (KPM14A, Koge Electricity, China) provide pressurised air at up to 64\,kPa and a combined maximum flow rate of 1.6\,L/min. Airflow to each of the four chambers is independently regulated by a high-speed miniature solenoid valve (S070; switching frequency ${>}100$\,Hz), with each pair of valves supplied by one pump. Each valve operates in binary switching mode; continuous pressure modulation was not implemented in the current system, as the actuator volume and valve response characteristics were optimised for rapid on/off switching. Motor drive for pumps and valves is managed by three DRV8833 dual H-bridge modules. System-level control is implemented on an ESP32-C3 mini microcontroller, which orchestrates pump speed (via PWM), valve timing, and wireless communication. A 3.7\,V, 1800\,mAh lithium-ion battery provides onboard power; at full system load ($\approx$4.3\,W), estimated battery life is approximately 1.5\,h, though the user study was conducted over approximately 4\,h without recharging, reflecting typical sub-maximal duty-cycle operation.

For actuator characterisation experiments requiring precise, independently regulated chamber pressure, a dedicated laboratory setup was employed. Four proportional pressure-control valves (ITV0010, SMC, Japan) supplied by a commercial air compressor (FIAC F6000/50) regulated each chamber independently, controlled via an ESP32 microcontroller.

\subsection*{Actuator Characterisation}

\textbf{Force--displacement characterisation.}
Single- and four-chamber actuators were evaluated on a universal testing machine (Alluris FMT-313, Germany). The compression platen descended at 20\,mm/min to the target height and then retracted (Supplementary Figure 1B). Chamber pressure was modulated using the proportional valve setup described above, and force and displacement data were sampled at 100\,Hz. Three repetitions were recorded per condition; results are reported as mean $\pm$ standard deviation.

\textbf{Frequency response characterisation.}
The dynamic test setup (Supplementary Figure~2C) consisted of a single-chamber actuator sandwiched between a uniaxial force sensor (S215, SMD Sensors, USA) and a rigid top constraint, with a replaceable spacer to precisely set the operating height (Supplementary Figure 1C--D). Chamber pressure was monitored by a miniature pressure sensor module (XGZP6847A, CFSensor, China). Frequency response was measured via a logarithmic frequency sweep from 1 to 100\,Hz (30 discrete frequencies; 10 steady-state cycles per frequency). The onset transient (first cycle) was discarded at each frequency, and the steady-state force amplitude was extracted using a lock-in detection algorithm referenced to the drive signal. Gain was normalised to the 1\,Hz reference value and expressed in decibels; the $-$3\,dB bandwidth was identified by interpolation.

\textbf{Step response.}
A pressure step command was applied and maintained for 3\,s before release. Rise time was defined as the interval for the force response to increase from 10\% to 90\% of the steady-state value; fall time was defined as the interval for decay from 90\% to 10\%.

\textbf{Durability testing.}
A sinusoidal pressure waveform (0--60\,kPa, 4\,s period) was applied to the pneumatic chamber continuously for 1{,}000 cycles using a solenoid valve, with output force and chamber pressure recorded simultaneously throughout.

\subsection*{Spatial Pressure Distribution Measurement}

A 6$\times$6 piezoresistive pressure distribution sensor array (FS-ARR-6X6-ROT, Legact, China) with 36 spatially distributed sensing elements was used to characterise the fingertip contact pressure profile. Element resistance decreases monotonically with applied normal force. The sensor was interposed between the haptic interface and a rigid backing surface (Supplementary Figure~2D). Raw sensor output (dimensionless values proportional to contact pressure) was reshaped into a 6$\times$6 spatial matrix and spatially interpolated using cubic spline interpolation to produce a smooth, continuous pressure map. A unified colour scale was applied across all conditions to enable direct cross-condition comparison.

\subsection*{VR Scenarios}

To demonstrate the haptic interface in interactive VR applications, three virtual scenes were designed corresponding to the three rendering modalities: contact configuration, directional sliding, and vibrotactile texture feedback. The system architecture is shown in Fig.~\ref{Fig-M-Subject1}A and Supplementary Figure~2E. The virtual reality environment was developed in Unity (Unity Technologies). Hand tracking was performed using an Ultraleap Leap Motion Controller in a desktop-mounted configuration, providing markerless optical tracking at approximately 120\,Hz. The fingertip of the index finger was modelled as a 30\,mm $\times$ 30\,mm square contact pad, intentionally oversized relative to the physical actuator footprint to improve detection robustness, subdivided into four equal quadrants spatially registered to the four pneumatic chambers. For each quadrant, the indentation depth was computed as the signed penetration distance between the quadrant corner and the nearest face of the virtual object's axis-aligned bounding box. Linear and angular velocities of the fingertip were derived from frame-to-frame differentiation of the tracked position and orientation, followed by exponential moving average (EMA) filtering ($\alpha = 0.15$ for linear velocity, $\alpha = 0.10$ for angular velocity) to suppress tracking noise. The computed haptic state was composed of four indentation values, four material identifiers, three velocity components, and the angular velocity. It was transmitted to the ESP32-based control box at 50\,Hz via UDP over a dedicated Wi-Fi link.

In the contact configuration scene, a virtual cube was placed on a platform. When the indentation of a given quadrant exceeded a threshold of 2\,mm, that region was classified as contacted and the corresponding pneumatic chamber was actuated. Different approach angles and edge contacts thus produced distinct spatial actuation patterns on the fingerpad, encoding the contact configuration through differential chamber activation. In the directional sliding scene, a large flat plate was placed in the virtual environment. Once the fingertip was in contact and the principal tangential velocity exceeded a threshold of 5\,mm/s, the system determined the dominant movement direction and initiated the corresponding sequential actuation pattern; the pattern continued until a different direction was detected or contact was lost. Rotational sliding was disabled in the current implementation. In the vibrotactile texture scene, three cubes with assigned surface materials (rough stone, medium-rough fabric, and smooth wood) were arranged side by side. When a quadrant entered contact with an object, the corresponding chamber was driven at the material-specific vibration frequency (5\,Hz for stone, 30\,Hz for fabric, 100\,Hz for wood), with all active chambers synchronised to a common oscillation phase to ensure perceptual coherence.

\subsection*{User Study Protocol}

Fifteen healthy adult participants (4 female, 11 male; age $30.4 \pm 3.7$\,years; height $173.3 \pm 11.5$\,cm; body mass $70.4 \pm 14.2$\,kg; index finger circumference $5.20 \pm 0.41$\,cm; see Supplementary Table~1) were enrolled. The experimental protocol was approved by the Ethics Committee of Scuola Superiore Sant'Anna (protocol 612025), and all procedures were conducted in accordance with the Declaration of Helsinki. Written informed consent was obtained from all participants prior to enrollment.

Participants wore the haptic device on the right index finger. Pneumatic actuation noise was masked using noise-cancelling headphones delivering white noise throughout the session (Supplementary Figure 2D). A custom MATLAB graphical interface delivered pseudo-randomized stimuli to the haptic device and logged participant responses and response times. Following stimulus onset, participants were permitted unlimited time to respond; once the experimenter logged the response, a 2\,s inter-stimulus interval elapsed before the subsequent trial began.

\textbf{Contact configuration task.}
Nine chamber actuation patterns (patterns 1--9; Fig.~\ref{Fig-M-Subject1}B) representing distinct fingertip contact configurations were presented. The active pattern was sustained until the participant's response was recorded.

\textbf{Directional sliding task.}
Six motion sequences were presented: four translational directions (left, right, up, down) and two rotational directions (clockwise, counterclockwise). Each translational stimulus consisted of a two-chamber sequential activation (100\,ms inter-chamber interval; 200\,ms actuation duration per chamber), repeated at 500\,ms rest intervals until the participant responded. Rotational stimuli involved sequential activation of three or four chambers following the same temporal structure.

\textbf{Vibrotactile frequency task.}
Stimuli were delivered at 5\,Hz, 30\,Hz, and 100\,Hz, representing low, mid, and high vibrotactile frequency ranges, respectively. Each stimulus was applied for 2\,s, followed by a 2\,s rest, repeating until the participant responded. Stimulus timing parameters were determined through pilot testing prior to the main study.

\textbf{Statistical analysis.}
For between-group comparisons (finger circumference and biological sex), data normality was first assessed using the Shapiro--Wilk test. Given non-normal distributions in the majority of measures, group differences were evaluated using the two-sided Mann--Whitney $U$ test. Classification accuracy for each task was compared against the corresponding chance level using a one-sample $t$-test. The significance threshold was set at $\alpha = 0.05$.

% ============================================================
\section*{Declaration statements}

\subsection*{Data Availability}

All data are available in the main text or the supplementary materials. The fabrication protocols, control software, and experimental datasets supporting the conclusions of this article are available from the corresponding author upon reasonable request.

\subsection*{Acknowledgments}

This project was funded by MSCA-DN / Project 101073374 - ReWIRE. Views and opinions expressed are however those of the author(s) only and do not necessarily reflect those of the European Union or the European Research Executive Agency (REA). Neither the European Union nor the granting authority can be held responsible for them.

\subsection*{Author Contributions}

Conceptualization: R. Chen, D. Leonardis, A. Frisoli.

Methodology and Investigation: R. Chen, D. Leonardis.

Experiments and Data Processing: R. Chen.

Writing---original draft: R. Chen.

Writing---review \& editing: R. Chen, D. Leonardis, A. Frisoli.

Funding acquisition and administration: A. Frisoli.

\subsection*{Conflicts of Interest}
The authors declare that there is no conflict of interest regarding the publication of this article.

\printbibliography

\end{document}